\documentclass[sigconf,screen,nonacm]{acmart}

\usepackage[utf8]{inputenc} 
\usepackage{hyperref}       
\usepackage{url}            
\usepackage{booktabs}       
\usepackage{amsfonts}       
\usepackage{nicefrac}       
\usepackage{microtype}      
\usepackage{xcolor}         
\usepackage{graphicx}
\usepackage{amsmath}
\usepackage{amsthm}
\usepackage{amsfonts}
\usepackage{wrapfig}
\usepackage{enumitem}
\usepackage{diagbox}
\usepackage{multirow}

\AtBeginDocument{%
  \providecommand\BibTeX{{%
    \normalfont B\kern-0.5em{\scshape i\kern-0.25em b}\kern-0.8em\TeX}}}

\newcommand{\hidecomment}[1]{}

\newcommand{\Rspace}{\mathbb R}
\newcommand{\Mspace}{\mathbb M}
\begin{document}

\title{Topological Representations of Local Explanations}

\author{Peter Xenopoulos}
\email{xenopoulos@nyu.edu}
\affiliation{%
  \institution{New York University}
  \city{New York}
  \country{NY}
}

\author{Gromit Chan}
\email{gromit.chan@nyu.edu}
\affiliation{%
  \institution{New York University}
  \city{New York}
  \country{NY}
}

\author{Harish Doraiswamy}
\email{harish.doraiswamy@microsoft.com}
\affiliation{%
  \institution{Microsoft Research}
  \city{Bengaluru}
  \country{India}
}

\author{Luis Gustavo Nonato}
\email{gnonato@icmc.usp.br}
\affiliation{%
  \institution{University of São Paulo}
  \city{São Paulo}
  \country{Brazil}
}

\author{Brian Barr}
\email{brian.barr@capitalone.com}
\affiliation{%
  \institution{Capital One}
  \city{New York}
  \country{NY}
}

\author{Claudio Silva}
\email{csilva@nyu.edu}
\affiliation{%
  \institution{New York University}
  \city{New York}
  \country{NY}
}

\renewcommand{\shortauthors}{}

\begin{abstract}
Local explainability methods -- those which seek to generate an explanation for each prediction -- are becoming increasingly prevalent due to the need for practitioners to rationalize their model outputs. However, comparing local explainability methods is difficult since they each generate outputs in various scales and dimensions. Furthermore, due to the stochastic nature of some explainability methods, it is possible for different runs of a method to produce contradictory explanations for a given observation. In this paper, we propose a topology-based framework to extract a simplified representation from a set of local explanations. We do so by first modeling the relationship between the explanation space and the model predictions as a scalar function. Then, we compute the topological skeleton of this function. This topological skeleton acts as a signature for such functions, which we use to compare different explanation methods. We demonstrate that our framework can not only reliably identify differences between explainability techniques but also provides stable representations. Then, we show how our framework can be used to identify appropriate parameters for local explainability methods. Our framework is simple, does not require complex optimizations, and can be broadly applied to most local explanation methods. We believe the practicality and versatility of our approach will help promote topology-based approaches as a tool for understanding and comparing explanation methods.
\end{abstract}




\maketitle

\section{Introduction} \label{sec:introduction}
Increasingly complex machine learning (ML) models are being deployed in industries such as healthcare, cybersecurity, and banking. While industries welcome the performance boost from these ML models, organizations are also starting to require their models to provide clear explanations for their predictions, as necessitated by regulations, such as GDPR~\cite{voigt2017eu}. Thus, explainable artificial intelligence (XAI) techniques are especially relevant, particularly those which provide explanations for individual samples. 

There are many techniques which provide \textit{local} explanations for predictive models~\cite{ribeiro2016should}. Given an input $x$ and predictive model $P$, a local explanation method typically generates a real-valued feature vector $e$ which represents the attributions of the input features to the predicted outcome. Since there are many local explanation methods available, it raises an important question -- \textit{how do we assess and compare local explainability frameworks}? One popular method is ablation testing~\cite{hooker2018benchmark}, which assesses the accuracy drops of an input by removing the important features identified by the explanation methods. However, such testing reveals only the ordering of features by their performance impact, not the structure of the explanation space. 

In this paper, we propose a simple but approach to globally assess local explanations (GALE). We accomplish this through an analysis of the topological properties of a given explanation space for binary classification problems. Specifically, we model an explanation method as a scalar function that captures the relationship between the explanation space and the class prediction. Using this function, we compute its topological skeleton. This skeleton is used to generate a topological signature which is then used to compare explanation methods. GALE is easy to implement and lacks complex optimizations or parameter tuning. We aim to address the following challenges arising from explanation technique assessments.

\begin{figure*}
    \centering
    \includegraphics[width=\textwidth]{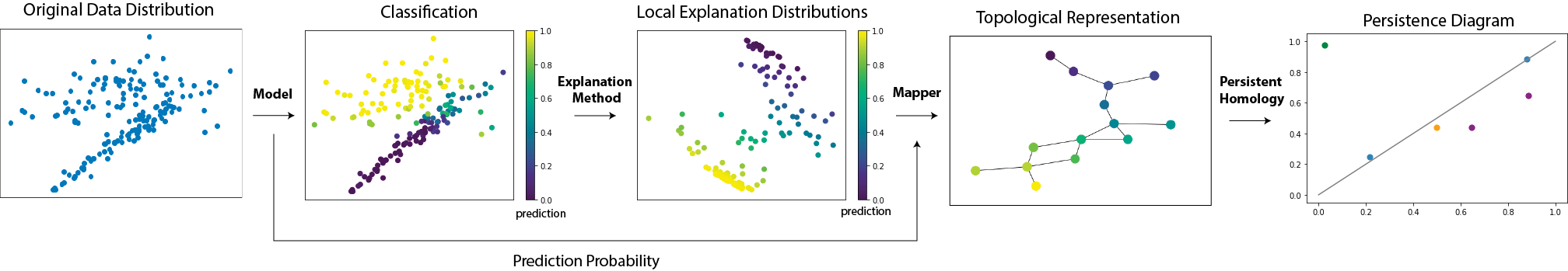}
    \caption{GALE is a framework which transforms local explanations of a data set into a topological representation which we can use to compare heterogenous explanation methods.}
    \label{fig:pipeline}
\end{figure*}

Prior to comparing multiple local explanation methods, it is important to first understand a given explanation method, and in particular, how it performs on any given data set. Even using a single explanation technique, it is common to observe different outcomes based on different parameters or simply from sampling variability. For example, explanations generated using identical parameters and inputs produced by Local Interpretable Model-Agnostic Explanations (LIME)~\cite{lime}, a popular local explainability method, can produce contradictory feature importance scores for the same input due to the stochastic nature of LIME. Furthermore, many of these explainability methods require user-specified parameters which can further impact the explanation output. While it is easy to observe the differences between multiple local explanations for a given input through visualization, such analysis is only useful for analyzing an explanation method for a single observation. Thus, a visual approach becomes cumbersome when applied to the entire data set. Moreover, it is almost impossible to understand how stable the explanation method is for that data. Doing so would necessitate generating a global measure or signature that can capture the structure of the explanation space with respect to the given data. Our aim is to compute a stable topological signature given an explanation method's output.

While comparing outputs from the same explanation method can be performed by measuring their distances (e.g., Euclidean distance), explanations generated from different methods generally cannot be directly compared since they have different value ranges and dimensions. For example, LIME and SHAP~\cite{shap}, another popular local explanation method, are on different scales. To compare two different explanation methods as a whole, a straightforward approach is to cluster the local explanations and compare the cluster similarities (e.g., Rand index) between them. However, there are two drawbacks to this approach. First, the values of explanations may not provide clear cluster structures, which makes comparisons using only on the explanation values difficult. Second, since explanations are inferred using the predictions of the ML model, the comparisons between different explanation methods are more faithfully measured by the relationships between the model predictions and the explanation output. As we show later, such properties are naturally captured by the shape, or more formally, the topological properties, of the explanation manifold. A crucial advantage of using topology for such an analysis is that it is agnostic to the range and dimensions of the explanation space and thus allows for direct comparison across different explanation methods.
To the best of our knowledge, our approach is the first to use computational topology for comparing XAI methods. We view topology as an intriguing direction for understanding and comparing explanation methods and that GALE, a powerful yet accessible framework, as the first step in that direction. The contributions of this work can be summarized as follows:
\begin{enumerate}[leftmargin=*]
\item We propose GALE (Global Assessing Local Explanations) a topology-based approach to generate a global signature for a given local explanation method's output. By providing a domain-agnostic signature for the explanation techniques, our approach allows comparison across heterogeneous explanation approaches.
\item We demonstrate that GALE is both stable and can elucidate differences between explanations through experiments on gradient-based and surrogate model techniques with real-world and synthetic data.
\item We show that GALE can be used to find appropriate parameters for local explanation methods by comparing topological signatures for different sets of parameters.
\end{enumerate}
\section{Related Work} \label{sec:related_work} 
When training models, practitioners can choose from a variety of model architectures, some of which are directly interpretable. For example, in generalized additive models (GAMs), feature attributions are additive and directly comparable from model to model. Chang~et~al.~\cite{DBLP:conf/kdd/ChangTLGC21} find that tree-based GAMs, such as explainable boosting machines~\cite{DBLP:conf/kdd/LouCGH13}, often display a balance of sparsity, fidelity and accuracy. Typically, GAMs are also easy to visualize and provide clear feature attributions to end-users. However, tree-based GAMs can be slow to train and may be outperformed by other models.

To balance the desire for machine learning model performance improvements along with growing calls for explainable model decisions, practitioners are increasingly turning to local, posthoc explainability methods~\cite{DBLP:journals/cacm/WeldB19}. These local explanation methods typically produce a vector representing the attributions of input features, and are generally well-received by users. Jeyakumar~et~al.~\cite{DBLP:conf/nips/JeyakumarNCGS20} find, through surveys with hundreds of non-technical users, that explanation-by-example and LIME were the preferred explanation styles. However, beyond user preferences, it is hard to compare explainability methods. Taxonomies provide some guidance on a unified language around explainability and raise important questions for explainability researchers to address~\cite{doshi2017towards}. Sokol and Flach propose \textit{Explainability Fact Sheets}, a taxonomy that compares explainability methods across their functional, operational, usability and safety properties~\cite{DBLP:conf/fat/SokolF20}. 

To assess and compare the attributions from local explanation methods, a straightforward way is to ablate the top K features ranked by the attributions and observe the decrease of the predicted output score. Varying the values of K and recording the output scores results in an ideally downward sloping output score curve. The lower the curve, the better the local explanation method since it shows that the explanation succeeds in identifying the important features. 

To avoid simply removing the top K features without considering the correlations among features, we can ablate the center of mass of the input instead~\cite{ghorbani2019interpretation}. Also, to avoid issues of model extrapolation on ablated inputs, we can retrain the model on the ablated data and measure the performance degradation~\cite{hooker2018benchmark}. Furthermore, local explanation methods can also be assessed by comparing their behavior between a randomly parameterized model and a trained model~\cite{adebayo2018sanity}. Besides ablation, we can measure the quality of explanations with metrics such as (in)fidelity and sensitivity under perturbations~\cite{yeh2019fidelity}, or impact score that measures the feature importance on decision marking process~\cite{lin2019explanations}. If the apriori of feature importance of the dataset~\cite{yang2019benchmarking} is known, the feature importance of input across different models can also be assessed. 


Topological data analysis (TDA) is beginning to be considered for explainability purposes. For example, Elhamdadi~et~al.~\cite{elhamdadi2021affectivetda} use TDA to study face poses used in affective computing and find that their topology-based approach captures known patterns. Van Veen proposed visual constraints on TDA output to aid interpretability~\cite{van2020novel}, specifically for viewing global, cluster and local explanations. However, there is little work on applying TDA to the outputs of explanation methods to assess the similarity between them.






\section{Methodology} \label{sec:methods}

In this section, we describe the topological concepts we use (Section~\ref{sec:topo-background}), our proposed approach to generate a global signature for local explanation methods  (Section~\ref{sec:our-method}), and our framework for parameter tuning (Section~\ref{sec:hyperparam-tuning}).

\subsection{Topological Background} \label{sec:topo-background}
\subsubsection{Reeb Graphs and the Mapper Algorithm}
\begin{figure}[t]
\centering
\includegraphics[width=\linewidth]{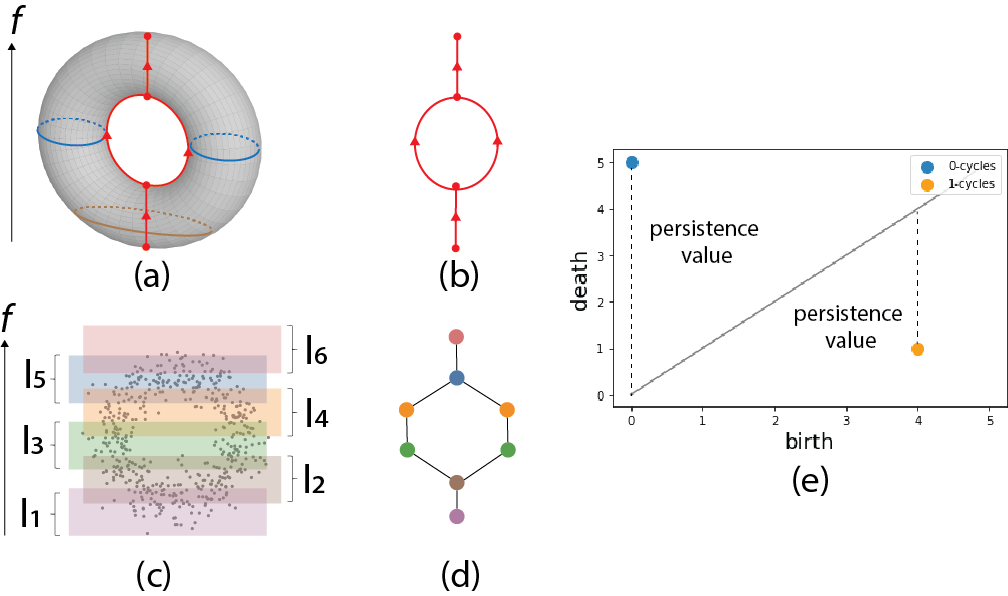}
\vspace{-0.1cm}
\caption{Approximate Reeb graph of a point cloud.
(a)~Height function defined on a torus. Level sets at two different heights are illustrated.
(b)~Reeb graph computed on the height function defined on the torus.
(c)~A point cloud sampled from the torus. 
(d)~The approximate Reeb graph computed using the Mapper algorithm when the height function is divided into 5 intervals as shown in (c).
}
\label{fig:reeb-graph}
\vspace{-0.3cm}
\end{figure}

Consider a scalar function $f:\Mspace \rightarrow \Rspace$, that maps points from a manifold $\Mspace$ to a real value.
The \textit{level set} $f^{-1}(a)$ at a given scalar value $a$ is the set of all points that have the function value $a$.
The \textit{Reeb graph}~\cite{reeb1946points} of $f$ is computed by contracting each of the connected components of the level sets of $f$ to a single point, resulting in a skeleton-like representation of the input.
Figure~\ref{fig:reeb-graph}(b) shows an example of the Reeb graph of the height function defined on a torus (Figure~\ref{fig:reeb-graph}(a)). 

However, many real-world data sets are available as sets of functions defined on a set of discrete high-dimensional points rather than continuous functions. 
The \textit{Mapper} algorithm~\cite{singh2007topological} computes an approximation of Reeb graph of some user-defined function (often called \textit{lens} or \textit{filter} function) of such data. 
Essentially, the Mapper algorithm divides the function range into a set of overlapping intervals and approximates the level sets to be the set of points that fall within each of these intervals.
The connected components of these approximate level sets are then computed by clustering the points that are part of a given interval.
Each cluster then forms a node of the approximate Reeb graph, and an edge is present between two nodes if they share one or more input points.

For example, consider the set of points in Figure~\ref{fig:reeb-graph}(c) that are sampled from the torus in Figure~\ref{fig:reeb-graph}(a). Assuming the height function, this is divided into a set of 5 intervals as shown in Figure~\ref{fig:reeb-graph}(c).
The connected components of the points that fall into these intervals are then used to generate the graph as shown in Figure~\ref{fig:reeb-graph}(d).
In this work, we will be using the Mapper algorithm to compute the topological graph to understand and compare the space defined by the different explanation methods.

\subsubsection{Topological Persistence}
Given a scalar value $a$, the sublevel set $f^{-1}((-\infty,a])$ is defined as the set of all points on the domain that have a function value less than or equal to $a$.
Consider a filtration of the input that sweeps the input scalar function $f$ with increasing function values. 
As the function value increases, the topology of the sublevel sets changes at the \textit{critical points} of the function (where its gradient is zero), and remains constant at other points. 
In particular, at a critical point, either a new topology is created, or some topology is destroyed. Here, topology is quantified by a class of $k$-dimensional cycles (or $k$-cycles). For example, a 0-dimensional cycle represents a connected component, a 1-dimensional cycle is a loop that represents a tunnel, and a 2-dimensional cycle bounds a void. 

A critical point is a creator if a new topology appears and a destroyer otherwise. 
Given a set of critical points $c_1,c_2,\ldots,c_m$, one can pair up each creator $c_i$ uniquely with a destroyer $c_j$ which destroys the topology created at $c_i$. We say that a topological feature is born at $c_j$ and it dies at $c_j$. The \textit{topological persistence}~\cite{edelsbrunner02} of this topological feature that is created at $c_i$ is defined as $f(c_j)-f(c_i)$, which intuitively indicates the lifetime of this feature in this sweep.

The above notion of persistence allows features to have an ``infinite" persistence, that is, there exist creators that are not paired with any destroyer. The notion of \textit{extended persistence}~\cite{edelsbrunner2010computational,AEHW06} extends the filtration to include a sweep over superlevel sets ($f^{-1}([a,\infty))$, thus allowing pairing of the above mentioned creators. Without loss of generality, we use the term persistence to mean extended persistence for the remainder of this paper.

Since we are working with functions defined over discrete points, we use the graph computed by the Mapper algorithm to compute the topological persistence of the features of the input. Here, the filtration is defined on the nodes and edges of the graph as follows. Each node is assigned a function value equal to the mean of the function values of the clustered points represented by that node. The order of the nodes added during the filtration (or sweep) is defined by the function value of the nodes. An edge is added during the step of the filtration as soon as both its endpoint nodes are added.
Note that the topological features, in this case, correspond to 0- and 1-cycles only. 
\subsubsection{Persistence Diagram}
A persistence diagram plots the topological features as a 2-dimensional scatter plot~\cite{DBLP:conf/focs/EdelsbrunnerLZ00}. Each point in the plot corresponds to a single feature and has $x$ and $y$ coordinates equal to its birth and death values respectively obtained from the extended filtration. 
The persistence value of each figure is then the height of the corresponding point above (or below) the line $x = y$.
Figure~\ref{fig:reeb-graph}(e) shows the persistence diagram computed using the Mapper graph.
Note that we plot all the features in the same plot, thus resulting in a single persistence diagram.

\begin{figure}
    \centering
    \includegraphics[width=\linewidth]{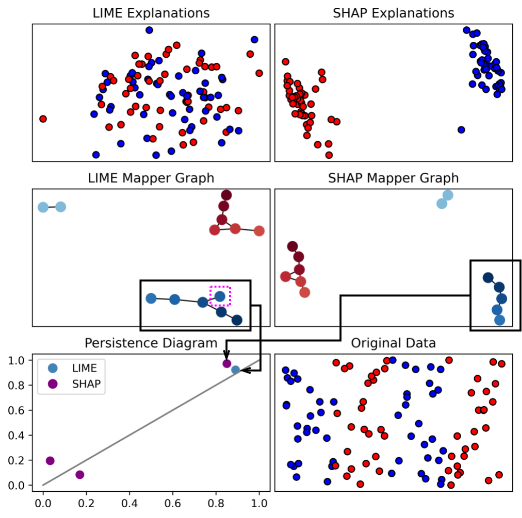}
    \caption{Comparing two sets of explanations using GALE. First, the explanation spaces are summarized using Mapper. From the Mapper output, we create persistence diagrams. Topological features correspond to points in the persistence diagram. In this example, we see that while the two Mapper outputs are generally similar, the Mapper output from the SHAP explanations is slightly different from that of the LIME explanations (see the node outlined in purple).}
    \label{fig:bottleneck}
\end{figure}

Persistence diagrams provide a useful mechanism to assess the structure of scalar functions. 
Moreover, it has also been shown that persistence diagrams are robust to noise~\cite{CEH07}. In other words, persistence diagrams are stable under small irregular perturbations in the data, and the distance between two such diagrams is bounded. We can also calculate the distance between two persistence diagrams. Hausdorff distance and bottleneck distance are two common measures used to compare two persistence diagrams. In this work, we use the \emph{bottleneck distance}~\cite{CEH07}, which pairs points between two persistence diagrams and then takes the maximum distance from this matching. In Figure~\ref{fig:bottleneck}, we show two explanation spaces generated by LIME and SHAP on a synthetic data set. It is clear that though the two spaces are quite different, and that a simple clustering (e.g., k-means) would return clusters with low cluster similarity, the Mapper outputs are very similar. We can see that in the persistence diagrams that the only features which differ are shown as the two points in the upper right, which have a small distance between each other.


\begin{figure*}
    \centering
    \includegraphics[width=\linewidth]{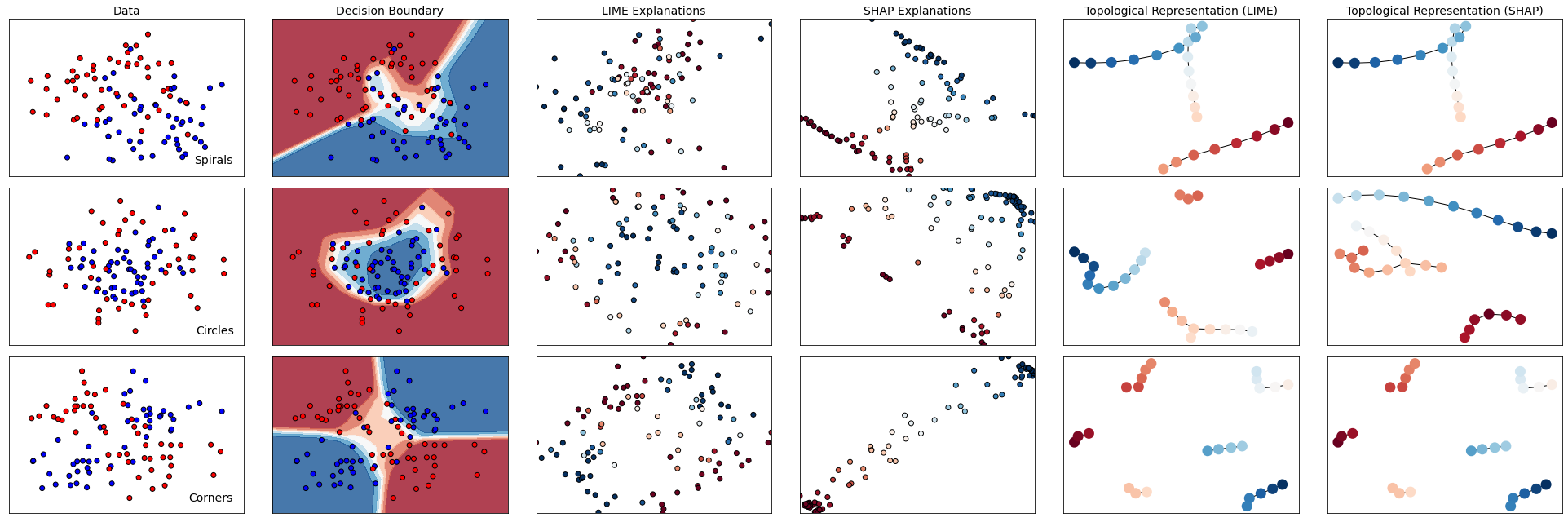}
    \caption{End-to-end pipeline showing the initial data, a trained model (decision boundary), LIME and SHAP explanations, and the topological representations for three different 2-dimensional synthetic classification problems with 100 observations each.}
    \label{fig:example_maps}
\end{figure*}

\subsection{GALE: Globally Assessing Local Explanations} \label{sec:our-method}
In this section, we introduce our approach to \textbf{g}lobally \textbf{a}ssesing \textbf{l}ocal \textbf{e}xplanations (GALE). Let $\mathcal{X}\subset\mathbb{R}^n$ be a data set and $P:\mathcal{X}\rightarrow [0,1]$ a binary classification model, where $P(x)\in[0,1]$ is the probability of $x\in\mathcal{X}$ belonging to the ``1'' class. Given $\mathcal{X}$ and the model $P$, 
a local explanation method can be seen as a mapping from the data set to the explanation space $E:\mathcal{X}\rightarrow\mathbb{R}^d$, where $E(x)$ provides the ``importance'' of the different attributes for the classification of $x\in\mathcal{X}$.
While $d=n$ for most approaches, $d$ can be greater than $n$ if the explanation method outputs more than a scalar value for each attribute of the input. 

The mapping $\mathcal{X}^\prime=E(\mathcal{X})$ gives rise to a point set in $\mathbb{R}^d$, which is the explanation manifold. 
We can define a function $f:\mathcal{X}^\prime\rightarrow [0,1]$ as
$f(x^\prime) = P(x)$, where $x^\prime=E(x), x\in\mathcal{X}$.
%
In GALE, we use the function $f$ as the lens function from which Mapper builds a summary representation $G$. Intuitively, $f$ captures the relationship between the explanations and the classification probabilities. From $G$, we can produce its persistence diagram $D$.

\begin{figure}
    \centering
    \includegraphics[width=\linewidth]{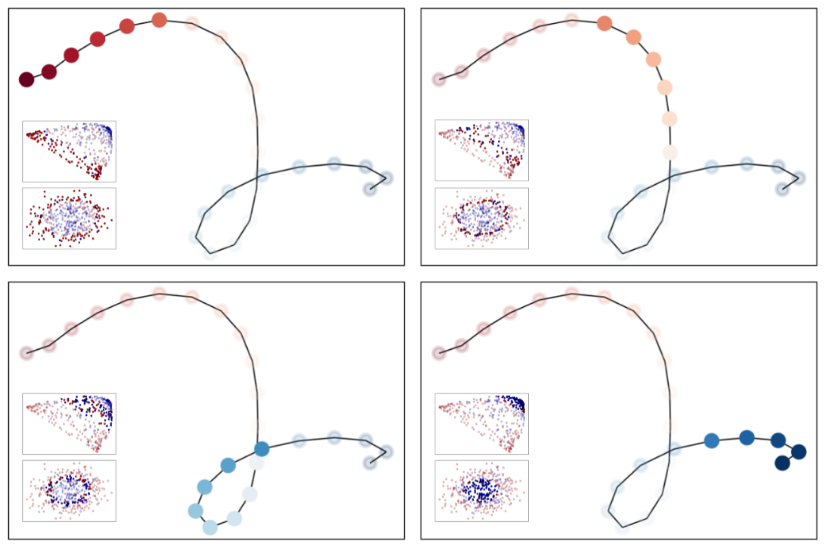}
    \caption{Mapper output for SHAP explanations on a synthetic circles data set with 400 observations. Each quadrant highlights one unique region of the graph, with the lower left of each quadrant showing the selected original data space (bottom) and explanation space (top).}
    \label{fig:graph_subsets}
\end{figure}

Since the explanation manifold can drastically vary across different methods and parameters, a direct comparison of the geometry of these manifolds is not possible. However, studying the topology of such functions allows us to analyze how explanation methods differ and thus, we can make geometry agnostic comparisons. To do so, we can take the distance between two persistence diagrams arising from two different explanation methods. If the distance is low, then the topologies of the explanation spaces are similar.
Furthermore, each topological feature (a point in the persistence diagram) can be easily mapped back to a set of input data points in $\mathcal{X}$, thus allowing us to also compare how the explanation space is spread across the input data. We show an example of GALE on three synthetic data sets in Figure~\ref{fig:example_maps}. In Figure~\ref{fig:graph_subsets}, we show how the Mapper graph corresponds to the original data and explanation spaces.

\subsection{Tuning MAPPER Parameters} \label{sec:hyperparam-tuning}
The Mapper algorithm used in this work requires three parameters:
1)~the \textit{resolution} $r$ of the lens function that defines the number of intervals into which the scalar function range is divided;
2)~the \textit{gain} $g$, which defines the percentage overlap between successive intervals; and
3)~the \textit{clustering} algorithm used, which may carry its own parameters.
%
The value of these parameters determines the structure of the resultant graph, and hence the persistence diagram. 

In an ideal scenario (e.g., in a dense point set input), increasing the resolution and decreasing the overlap would result in the graph computed using the Mapper algorithm converging to the Reeb graph. However, in real-world data, too high a resolution or too low a gain can result in the graph being a set of disconnected nodes. In our implementation, we use agglomerative clustering, as implemented in the \texttt{sklearn} Python library, which recursively merges pair of clusters. We consider the ``distance threshold'' parameter, which is the linkage distance threshold, above which clusters are not merged. Due to the varying spaces arising from different explanation methods, we set the distance threshold parameter parameter to a fraction of the maximum minus the minimum value of the explanation space. 

Depending on the point distribution, small changes in the parameters can drastically change the resultant graph, and hence the persistence diagram. A trivial option would be to use a resolution of one, which would simply be equal to clustering the points. Poor selections of Mapper parameters can produce graphs which are disjointed or unstable under small perturbations to the input. Blaser and Aupetit~\cite{blaser2020research} suggest that direct measures for measuring the quality of Mapper output could be a combination of cluster quality and their consistency, however, they note that they found no previous work in this direction.

We are therefore interested in identifying a set of Mapper parameters that not only produces a ``stable'' graph computation but also produces graphs with clear structure (i.e., a small number of connected components.) When plotting the Mapper graphs, we use the Fruchterman-Reingold force-directed algorithm~\cite{DBLP:journals/spe/FruchtermanR91}.
%

To measure the aforementioned stability, we use bootstrapping~\cite{chazal2017robust} to compute the confidence intervals for the generated graph in terms of the bottleneck distances. Consider a set of input explanation values $E = \{e_1,e_2,...,e_n\}$ and the persistence diagram $D$ generated by its Mapper graph $G$. $G$ is generated from some parameter set $H$, which contains the resolution, gain and agglomerative clustering distance threshold. Then, for each iteration in the bootstrap, we sample with replacement from $E$ to construct $E^* = \{e^*_1,e^*_2,...,e^*_n\}$, compute Mapper graph $G^*$ and persistence diagram $D^*$ for this input, and calculate the bottleneck distance $d_b(D,D^*)$. Using the distribution of distances created by these iterations, we can then find the value $\hat{b}_\alpha$ such that

\begin{equation}
    P(d_b(D,D^*) \geq \hat{b}_\alpha) = \alpha
\end{equation}

A parameter set which is stable will produce a low value for $\hat{b}_\alpha$. Additionally, for each iteration, we count the number of connected components in $G^*$. We can find the value $\hat{c}_\alpha$ such that 
\begin{equation}
    P(\text{connected components of } G^* \geq \hat{c}_\alpha) = \alpha
\end{equation}

To tune the Mapper parameters, we perform a grid search over the resolution, gain and distance threshold parameters. For each parameter combination, we use 100 bootstrap iterations to estimate $\hat{b}_\alpha$ and $\hat{c}_\alpha$ for $\alpha = 0.05$. Using the estimated stability and connected components for every combination of parameters, we then iterate through the parameter set and greedily select the best combination. While this approach does not guarantee an optimal solution (e.g., we could arrive at a parameter set which has a low $\hat{b}_\alpha$ but a high $\hat{c}_\alpha$), we find that a greedy strategy works well and led to interpretable results on a wide array of synthetic and real world data.


\section{Evaluation} \label{sec:results}
In this section, we evaluate GALE by demonstrating that GALE (1) can be used to compare local explanations, (2) generates stable representations and (3) can be used to tune parameters for popular local explanation methods.

\subsection{Comparing Local Explanation Methods} \label{sec:comparison}
\subsubsection{Comparing Baselines in Gradient-Based Methods}
Here, we illustrate how GALE can be used to understand and compare different explanation methods. In particular, we aim to address a common challenge in using local explanation techniques---what should be the baseline for gradient-based explanation methods?
Baselines act as references to compare the relative importance of features for an input so that attributions can be calculated. We provide examples with both synthetic and real data sets to compare the behavior between different baselines. To be specific, we apply three explanation methods---Integrated Gradients~\cite{sundararajan2017axiomatic}, DeepLIFT~\cite{shrikumar2017learning}, and SHAP~\cite{AEHW06} with five different baselines: 
(1)~zero baseline (an input with all values being zeros), 
(2)~maximum distance baseline~\cite{sturmfels2020visualizing}, 
(3)~Gaussian baseline~\cite{smilkov2017smoothgrad, sturmfels2020visualizing}, 
(4)~uniform baseline~\cite{sturmfels2020visualizing}, and 
(5)~a trained baseline~\cite{izzo2020baseline}, 
resulting in 15 tables of explanations for each experiment.

Our goal in this experiment is to illustrate how GALE can be used to identify explanation methods that behave differently. To do so, we generate a synthetic data set with 5 features and determine the labels with an extremely simple logic---the input will be labeled as ``1'' if any of the columns contain a ``zero'' as value. Otherwise, they are labeled as ``0''. When working with real data sets, it is normal for them to contain zero values that may provide important information. Under our construction, only zero values are important to the classification. Thus we expect the zero baseline is the only reference that is not a neutral input to the classifier. We created 20 different synthetic data sets under the aforementioned conditions, where each set varied in its rate of ``0'' labeled instances.

\begin{figure}
    \centering
    \includegraphics[width=\linewidth]{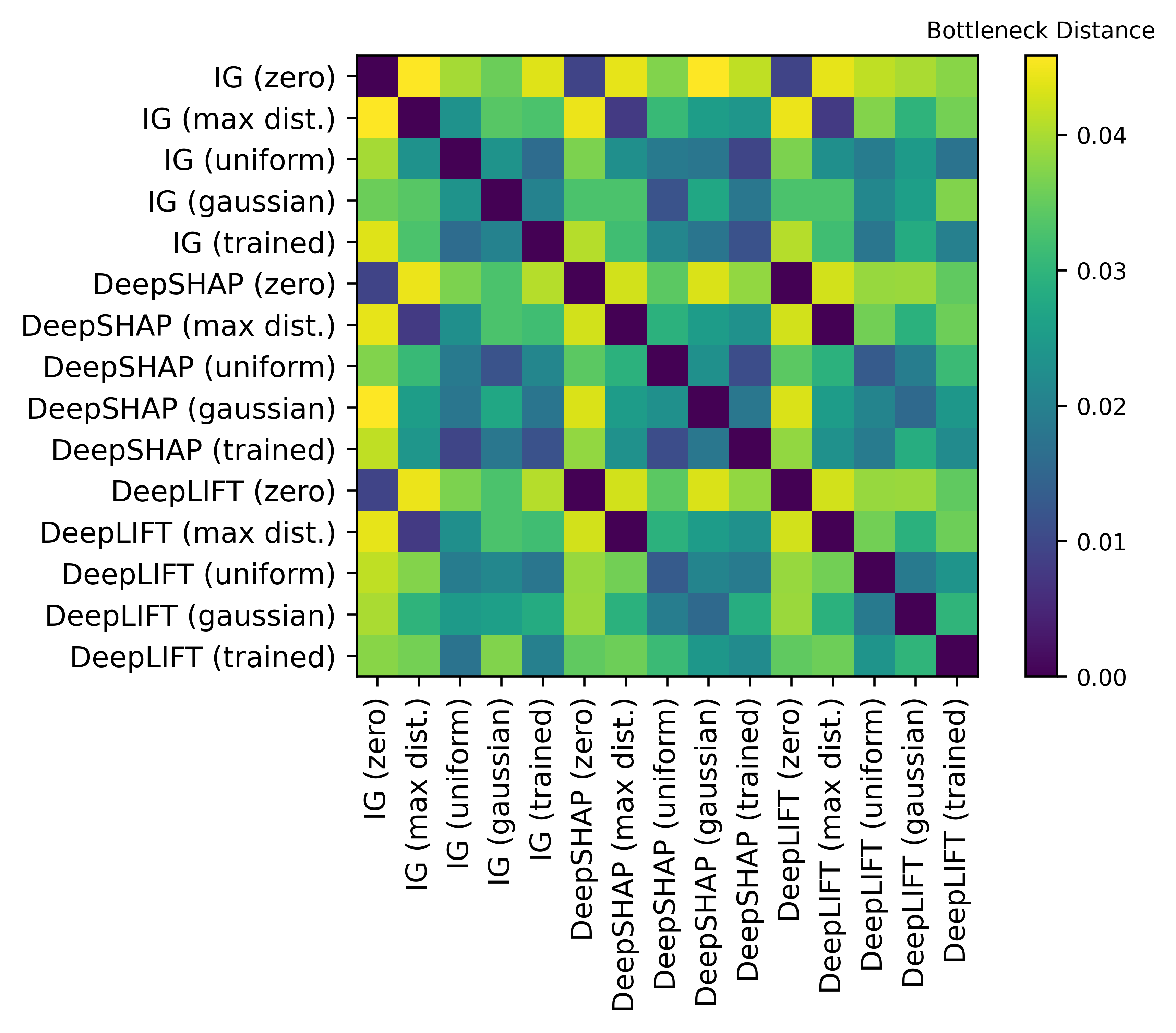}
    \caption{Pairwise distance matrix showing the average bottleneck differences among gradient-based explanation methods across 20 synthetic datasets. We see that the methods which use the zero-baseline have higher bottleneck distances to the explanation topologies using other baselines.}
    \label{fig:gradient_based_matrix}
\end{figure}

We first calculate a pairwise distance matrix for each of the 20 data sets. Then, we average the distance matrices, and show the result in Figure~\ref{fig:gradient_based_matrix}. We can see that there are three rows and columns which have much brighter colors (i.e. greater average distances) than the others, meaning that these three explanations differ greatly as compared to the other explanations. Such observation is consistent with the fact that zero baselines produce different feature attributions by treating the important values as neutral references. In general, any set of attributions having a significant deviation from the consensus should be treated with caution.

\begin{figure*}
    \centering
    \includegraphics[width=\linewidth]{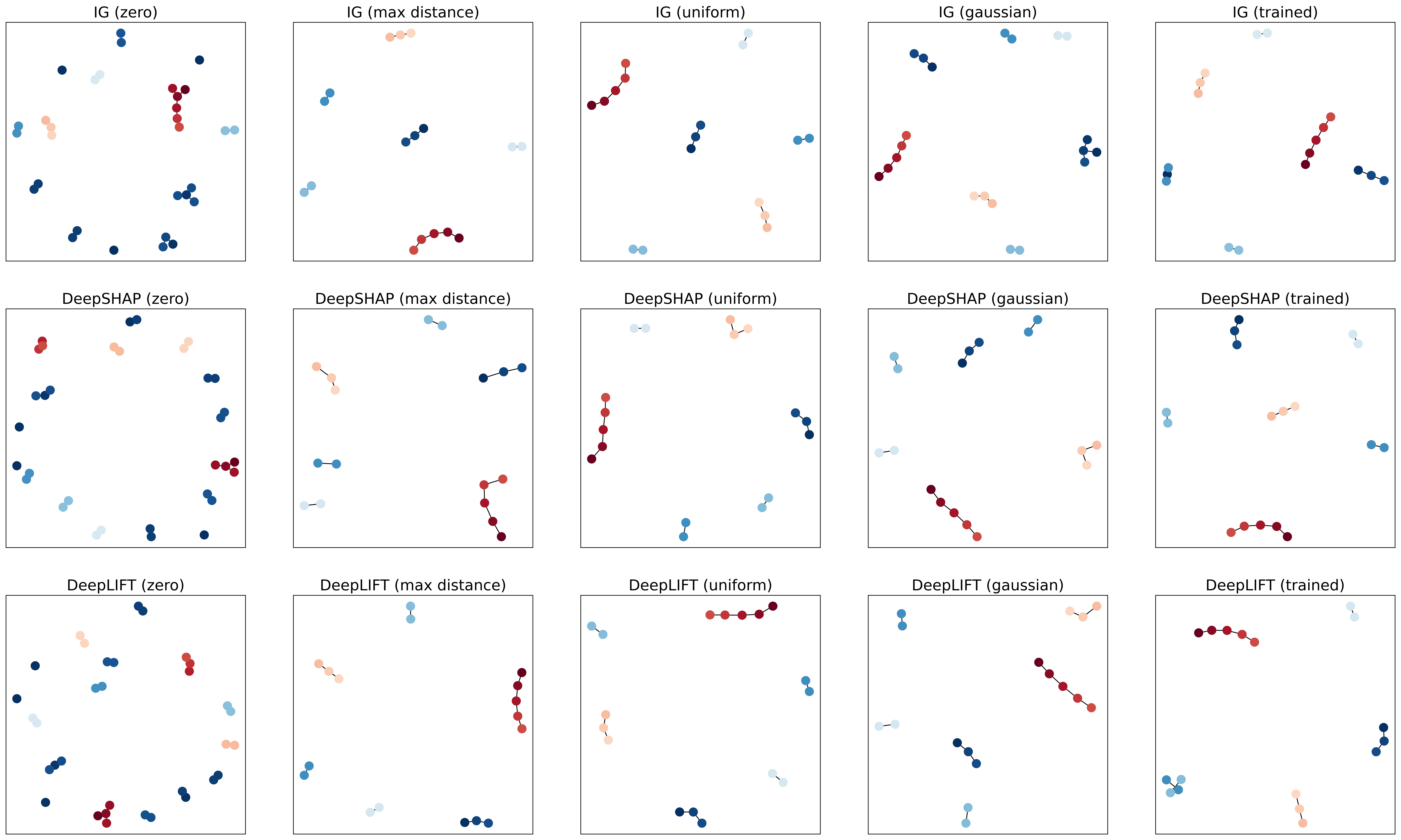}
    \caption{Visualizing the differences between the generated topologies reveals that the zero-baseline topologies contain more, smaller connected components than those from other explanation methods.}
    \label{fig:gradient_based_graphs}
\end{figure*}

When we explore the corresponding topological skeletons, we observe that these explanations typically have a more scattered graph compared with the others. Such visual evidence provide a sense that the explanation values differ greatly among the inputs in the data set. This evidence can be found statistically as well. When computing the variances of the explanation values among different outcomes, the variances are generally high among the outputs with scattered topological graphs. We show an example in Figure~\ref{fig:gradient_based_graphs}. Therefore, the topological representations also provide an overview of the variances of the explanations.

\subsubsection{Comparing LIME, SHAP and Explainable Boosting Machines}
Many local explainability methods are run post-hoc, meaning that first a model is trained and subsequently an explainability method is applied to the model's output to produce explanations. However, understanding the computed explanation space is oftentimes unintuitive. Generalized additive models, or GAMs, are a modeling approach whereby each feature contributes additively to a model's prediction. One desirable trait of this approach is that the feature contributions are easily interpretable, and thus present a alternative to LIME and SHAP. Explainable Boosting Machine (EBM) are a popular interpretable tree-based GAM from the \texttt{interpret} Python library~\cite{nori2019interpretml}. While we expect there to be consensus between LIME and Kernel SHAP explanations, since the two methods are similar~\cite{lundberg2017unified}, it is unknown how the explanations from these methods compare to those from EBMs.

We consider both synthetic and real world data sets for our experiments. For our synthetic data sets, we consider the spirals, circles and corners data from Figure~\ref{fig:example_maps}, one linearly-separable 2-dimensional data set, three ``toy'' data sets and the Pima Indians diabetes~\cite{smith1988using} data set. In the toy data sets, we have six features $\{x_{1}, ..., x_{6}\}$, and the target for observation $i$ is determined by $y_i = x_{i,1} + x_{i,2} + x_{i,3} + x_{i,4}$ and each feature is independent from one another. In ``independent'', each feature is weighted equally and is independent from one another. In ``flip'', $y_i = x_{i,1} - x_{i,2} + x_{i,3} - x_{i,4}$. Finally, in ``interaction'', we add equal-weighted interactions in the form of $y_i = \sum_{k=1}^4 x_{i,k} + \sum_{j=2}^4 10 x_{i,1} \cdot x_{i,j}$. For each data set, we train a 2-layer feedforward neural network with 64 hidden units per layer, as well as an Explainable Boosting Machine (EBM). We use the max number of features and a 50-sample neighborhood for our LIME explainer. 

\begin{table*}[]
\centering
\begin{tabular}{|c|c|c|c|c|c|c|}
\hline
\textbf{Data} & \textbf{Type} & \textbf{Obervations} & \textbf{Features} & \backslashbox{\textbf{LIME}}{\textbf{SHAP}} & \backslashbox{\textbf{LIME}}{\textbf{EBM}} & \backslashbox{\textbf{SHAP}}{\textbf{EBM}} \\ \hline
Spirals            & Synthetic & 100   & 2  & 0.00 & 0.11 & 0.11 \\ \hline
Circles            & Synthetic & 100   & 2  & 0.12 & 0.34 & 0.35 \\ \hline
Corners            & Synthetic & 100   & 2  & 0.08 & 0.15 & 0.15 \\ \hline
Linearly Separable & Synthetic & 100   & 2  & 0.00 & 0.02 & 0.02 \\ \hline
Toy                & Synthetic & 300   & 6  & 0.00 & 0.03 & 0.03 \\ \hline
Toy-Flip           & Synthetic & 300   & 6  & 0.00 & 0.08 & 0.08 \\ \hline
Toy-Interaction    & Synthetic & 300   & 6  & 0.01 & 0.11 & 0.11 \\ \hline
Breast Cancer      & Real      & 569   & 30 & 0.00 & 0.02 & 0.02 \\ \hline
Diabetes           & Real      & 768   & 8  & 0.00 & 0.45 & 0.45 \\ \hline
Sonar              & Real      & 208   & 60 & 0.00 & 0.00 & 0.00 \\ \hline
Banknote           & Real      & 1,372 & 4  & 0.00 & 0.00 & 0.00 \\ \hline
Ionosphere         & Real      & 351   & 34 & 0.00 & 0.01 & 0.01 \\ \hline
Bank Marketing     & Real      & 4,119 & 19 & 0.01 & 0.44 & 0.44 \\ \hline
German Credit      & Real      & 1,000 & 24 & 0.00 & 0.45 & 0.45 \\ \hline
\end{tabular}
\caption{Bottleneck distances between LIME, SHAP and EBM explanation topologies for synthetic and real world data. Intuitively, LIME and SHAP generally have low-bottleneck distances, indicating that their explanation topologies are similar.}
\label{tab:lime-shap-ebm}
\end{table*}

We show the bottleneck distances between the explanations from LIME, SHAP and EBM on each data set in Table~\ref{tab:lime-shap-ebm}. As expected, we see strong consensus between LIME and SHAP explanations, particularly in our real world data sets. However, we see varied consensus between EBM and LIME/SHAP explanations. Noticeably, there was more consensus among the three explanation methods when the prediction problem was ``easy'', such as in the linearly separable and toy synthetic data sets. One possible explanation for the lack of consensus on some data sets is that EBMs, unlike LIME and SHAP, for a $k$-dimensional data set, return explanations of size $j$, where $j \geq k$. This is because EBMs include an interaction component, where the interactions are detected by the EBM algorithm. We can enforce zero interactions when training, so that $j = k$ in the explanation output. However, by enforcing no feature interactions, the effect on explanation consensus was ambiguous, which suggests that the topologies generated by GAMs and LIME/SHAP can differ significantly.



\subsection{Evaluating Topological Stability}
Although the previous method sought to determine appropriate parameters for a local explainability method, many methods, like LIME, rely on stochastic behavior, which can produce different explanations for different runs, even when given the same parameters. It is important that our topological representation is resistant to sampling variability incurred by local explanation methods. In this section, we benchmark GALE's stability with regards to the variation induced by local explainability methods. Additionally, we show how our greedy approach to Mapper parameter selection aids in finding stable topological representations.

To determine the robustness of GALE to randomness induced by a local explanation method, for each synthetic data set and the diabetes data set, we run LIME 30 times. We restricted our experiments to these data sets for computational purposes. For each run, we generate two Mapper outputs: one which uses the Mapper parameters found via a greedy search, as described in section~\ref{sec:our-method}, and one which uses fixed parameters: a resolution of 15, a gain of 0.3 and an agglomerative distance threshold of 0.3 times the range of the explanation space. We then generate two bottleneck distance matrices -- one for distances between the optimized Mappers and another for distances between the fixed parameter Mappers. In Table~\ref{tab:stability}, we report the average row-sum for each of these matrices for each synthetic data set, as well as the average number of connected components from the mappers generated. The higher the row-sum, the less consensus there is among the explanation topologies produced by Mapper. We see that using our greedy Mapper parameter search procedure, our output is resistant to variation in LIME output. Although for some data sets, the fixed parameters returned a lower row-wise sum, we saw that in these cases, the fixed parameters also returned a higher number of connected components.

\begin{table}[]
\centering
\begin{tabular}{|c|cc|cc|}
\hline
\multirow{2}{*}{\textbf{Data}} &
  \multicolumn{2}{c|}{\textbf{Row-Wise Distance}} &
  \multicolumn{2}{c|}{\textbf{\begin{tabular}[c]{@{}c@{}}Connected \\ Components\end{tabular}}} \\ \cline{2-5} 
 &
  \multicolumn{1}{c|}{\textit{Greedy}} &
  \textit{Fixed} &
  \multicolumn{1}{c|}{\textit{Greedy}} &
  \textit{Fixed} \\ \hline
Spirals            & \multicolumn{1}{c|}{0.00} & 0.83 & \multicolumn{1}{c|}{2.0} & 4.1  \\ \hline
Circles            & \multicolumn{1}{c|}{1.88} & 2.56 & \multicolumn{1}{c|}{6.1} & 10.5 \\ \hline
Corners            & \multicolumn{1}{c|}{1.22} & 1.61 & \multicolumn{1}{c|}{9.0} & 14.6 \\ \hline
Linearly Separable & \multicolumn{1}{c|}{0.00} & 0.00 & \multicolumn{1}{c|}{4.0} & 4.0  \\ \hline
Toy                & \multicolumn{1}{c|}{0.03} & 0.00 & \multicolumn{1}{c|}{2.0} & 3.0  \\ \hline
Toy-Flip           & \multicolumn{1}{c|}{0.00} & 0.00 & \multicolumn{1}{c|}{2.0} & 2.2  \\ \hline
Toy-Interaction    & \multicolumn{1}{c|}{0.16} & 0.00 & \multicolumn{1}{c|}{2.0} & 5.0  \\ \hline
Diabetes           & \multicolumn{1}{c|}{0.38} & 2.70 & \multicolumn{1}{c|}{4.3} & 36.4 \\ \hline
\end{tabular}
\caption{Average row-wise sum of distance matrix and average connected component counts after using parameters found by our greedy parameter search. We find that our parameter tuning procedure produces more connected and stable Mapper output.}
\label{tab:stability}
\end{table}

\subsection{Using GALE to Tune Explainability Method Parameters} \label{sec:tuning-explain-params}
Many local explainability methods use a variety of parameters which guide the output. For example, two important parameters in LIME are the maximum number of features present in the explanation and the size of the neighborhood to learn the linear model. Likewise, in SHAP, depending on the explanation method, one can specify parameters such as the number of times to re-evaluate the model when explaining each prediction in Kernel SHAP, or the limit on the number of trees used in Tree SHAP. Choosing an appropriate set of parameters is important, as there can be significant variability in explanations coming from even just slightly different parameters~\cite{visani2020statistical}. After establishing that GALE can not only elucidate the differences between explanation sets but is also stable, we now show how to use GALE to tune the parameters of a local explainability methods, in particular, LIME. We consider the toy synthetic data, where each feature is independent of one another, used in Section~\ref{sec:comparison} for our experiment. We train a random forest classifier on this data using the default parameters from \texttt{sklearn}. Our goal is to determine the number of features to use for our LIME explanation, which we know to be four. We set the size of the neighborhood to 50 observations.

\begin{figure}
    \centering
    \includegraphics[width=\linewidth]{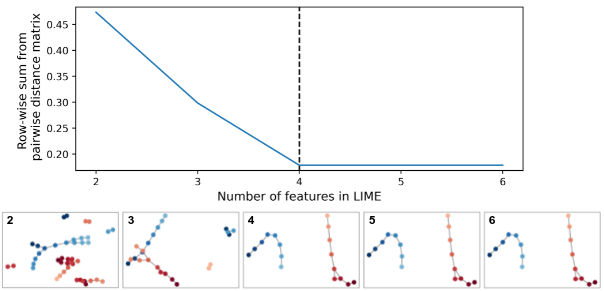}
    \caption{Row-wise sum of distance matrix for Mapper graphs computed on LIME explanations of different feature counts. Since the average row-wise distance stays constant when using 4 or more features for LIME, we know that the produced topologies are similar for these parameter values.}
    \label{fig:lime_feature_selection}
\end{figure}

Using our Mapper parameter tuning technique to find appropriate parameters, we compute the corresponding graph for explanations using $k$ features, where $k \in \{2, ..., 6\}$. Then, we calculate the bottleneck distance between each of the persistence diagrams to produce a distance matrix. Next, we calculate the row-wise sum of this distance matrix. Persistence diagrams which differ strongly from others will have a high row-wise sum. We plot the row-wise sum in Figure~\ref{fig:lime_feature_selection}. We see that when the number of features is four or higher, the returned topological signatures are identical. Thus, one may conclude that the addition of a fifth or sixth feature in the LIME explanations has no effect on the overall topology of the explanations. Likewise, by only including two or three features in LIME, one may not be estimating appropriate explanations.

\section{Discussion} \label{sec:discussion}
In this work, we outline and evaluate a methodology to compare local explainability methods through topological data analysis (TDA). While TDA shows promise in tackling challenges encountered in XAI applications, there are still several shortcomings which we plan to address in the future. The main constraint on our current methodology is that we focus on binary classification problems using tabular data. Furthermore, a deeper understanding of the effect and tuning of the Mapper parameters is also warranted.

Extending GALE to handle multi-class classifiers poses an important challenge---modeling a function similar to what is being done results in a multi-variate function, for which defining the appropriate filtration to compute the persistence diagrams becomes non-trivial. One possibility is to transform multi-class probabilities into scalar values. Additionally, future work should be directed towards extending GALE to accommodate prediction problems in the image and sequence spaces. A simple approach could be to reshape image explanations into a vector, and then use the approach we outlined here. While we did not evaluate GALE on regression problems, the evaluation would be similar, except for changing the lens function from the class predictions to the regression output.

Lastly, the output from Mapper is sensitive to the algorithm's parameters. While we introduced a greedy approach to find a good set of parameters which give stable and connected representations, more work is needed to develop alternative ways to tune the parameters. Furthermore, we only considered agglomerative clustering with single linkage. Within agglomerative clustering, there are a variety of linkage criterion to consider, such as Ward or average linkage. Additionally, one may consider other clustering algorithms, such as DBSCAN~\cite{DBLP:journals/tods/SchubertSEKX17}.
\section{Conclusion} \label{sec:conclusion}
In this work, we present GALE, a topology-based framework to globally summarize the outputs of local explanation methods. To do so, we compute a topological skeleton of a scalar function that captures the relationship between the explanation space, such as those created by LIME or SHAP, and a model's predictions. From this skeleton, we compute persistence diagrams. Then, to compare explanations, we compare persistence diagrams. The benefits of our method are that GALE (1) allows for easy comparison of heterogenous local explanations, differing in dimensionality or feature values, (2) is resistant to outliers and variation induced by local explanation method sampling procedures, and (3) can be used to optimize hyperparameters in local explanation methods. We validate our approach through experiments on both real and synthetic data sets. GALE is simple, lacks complex optimizations, and can be broadly applied to many local explanation methods. We believe our method is a fruitful and accessible first step in using topological techniques to understand explainability methods.

\begin{acks}
We would like to thank Bruno Coelho and Richen Du for their support in running preliminary experiments. This collaboration has been supported by a grant from Capital One. Silva's research has also been supported by NASA; NSF awards CNS-1229185, CCF-1533564, CNS-1544753, CNS-1730396, CNS-1828576, CNS-1626098; and DARPA. Any opinions, findings, and conclusions or recommendations expressed in this material are those of the authors and do not necessarily reflect the views of DARPA, NSF, NASA, or Capital One. 
\end{acks}

\bibliographystyle{ACM-Reference-Format}
\bibliography{00_bibliography}


\end{document}